\documentclass[runningheads]{llncs}
\usepackage[T1]{fontenc}
\usepackage{graphicx}
\usepackage{hyperref}
\usepackage{multirow}
\usepackage{array}
\usepackage{threeparttable}
\usepackage{amsfonts}
\usepackage{amsmath}
\newcolumntype{C}[1]{>{\centering\let\newline\\\arraybackslash\hspace{0pt}}m{#1}}
\usepackage{color}

\begin{document}
\title{Explanation-based Counterfactual Retraining(XCR): A Calibration Method for Black-box Models}
\author{Liu Zhendong\inst{1} \and
Wenyu Jiang\inst{2} \and
Yi Zhang\inst{3} \and Chongjun Wang\inst{*}}
\institute{Nanjing University, China} %\and
%Springer Heidelberg, Tiergartenstr. 17, 69121 Heidelberg, Germany

%
\maketitle              % typeset the header of the contribution
\begin{abstract}
  With the rapid development of eXplainable Artificial Intelligence (XAI), a long line of past
  papers have shown 
  concerns about the Out-of-Distribution (OOD) problem in perturbation-based post-hoc XAI models
   and explanations are socially misaligned. We explore the limitations of post-hoc explanation 
   methods that use approximators to mimic the behavior of black-box models. Then we propose eXplanation-based Counterfactual Retraining (XCR), which extracts feature importance fastly. XCR applies the explanations generated by the XAI model as counterfactual input to retrain the black-box model to address OOD and social misalignment problems. Evaluation of popular image datasets shows that XCR can even improve model performance when retaining 12.5\% of the most crucial features without changing the black-box model structure.
  Furthermore, the evaluation of the benchmark of corruption datasets shows that 
  the XCR is very helpful for improving model robustness and positively impacts 
  the calibration of OOD problems. Even though not calibrated in the validation set 
  like some OOD calibration methods, the corrupted data metric outperforms existing methods. 
  Our method also beats current OOD calibration methods on the OOD calibration metric 
  if calibration on the validation set is applied.
\end{abstract}
\section{Introduction}
Artificial intelligence (AI) has developed rapidly in recent years and has
 been widely used in many fields, such as natural language processing, computer vision, etc.
  The performance of deep neural networks (DNNs) has significant advantages over traditional 
  machine learning algorithms. However, DNNs are a typical black-box mothed. It is difficult 
  for people to understand their internal behavior, limiting the application of the DNNs in 
  fields with high security and robustness requirements, such as medical diagnosis, financial 
  risk control, etc. In order to solve the problem that most of the current AI models are black-box
   models, researchers proposed the concept of eXplainable Artificial Intelligence (XAI). With the
    development of AI, some explainable and privacy-safe requirements have been raised, such 
    as the General Data Protection Regulation (GDPR) proposed by the European Parliament, which
     stipulates that everyone has the right to have a meaningful understanding of the logic involved
      in the processing of personal data\cite{li2020survey}. Model-agnostic methods are important 
      in current XAI solutions, i.e., explanation methods can be broadly used for any 
      black-box model. Among model-agnostic explanation methods, perturbation-based explanation 
      plays a key role. Perturbation-based algorithms focus on the relationship between different 
      perturbed inputs and model outputs, e.g., 
      methods such as OCCLUSION\cite{zeiler2014visualizing}, RISE\cite{petsiuk2018rise}, 
      Extremal Perturbations\cite{fong2019understanding}.\\
      \indent
      However, with the development of AI, concerns about OOD issues have gradually emerged. Due to the influence of OOD data, the overconfidence of the model, low robustness, 
      and poor performance in the open environment are exposed. In the perturbation-based  XAI method, it is generally necessary to add perturbation to the original input or remove some features. 
      For example, in the RISE\cite{petsiuk2018rise} method, the input needs to be randomly masked 
      and then input into the black-box model. Some researchers \cite{hooker2019benchmark} discussed 
      whether there is an OOD problem in the output of the black-box model based on the perturbation 
      method. The generation of explanations may not be faithful to the original model. 
      The explanations are not generated to be generated by perturbations. 
      It may be caused by a shift in the input data distribution. In addition, 
      perturbation-based explanation usually requires multiple perturbations and thousands of neural
       networks forward passes for a single raw input. The results are averaged 
       to obtain the feature importance of the input. For example, the RISE\cite{petsiuk2018rise} 
       method uses thousands of random masks to get explanations. This way of generating 
       feature importance explanations is computationally expensive and is not conducive
        to the deployment of real-time models.\\
        \indent
        Hase et al.\cite{hase2021out} proposed that explanations generated using a standardly trained neural model are 
        usually socially misaligned, which is a concept introduced initially by \cite{jacovi2021aligning}. The training process and post-hoc feature
         importance explanation of classical neural networks are socially misaligned. Simply put, 
         we want a feature importance explanation to choose the information on which the model makes a decision, not the information picked out after the decision has been made. 
         In addition, we believe that through the study of explainable artificial intelligence, we can deeply understand the details of the model,
          obtain ideas and methods to improve the model's performance, open the black box system of 
          the deep learning model, and enhance the interactivity of the model. However, post-hoc 
          explanation generally does not improve model performance but only explains existing models and even damages the performance of neural network models.
           Some researchers\cite{cui2019chip} believe that there is a trade-off relationship
            between model explainability and model performance, which is contrary to the research
             purpose of XAI that we firmly believe in improving model performance by understanding model mechanism.\\
             \indent
             This paper discusses the limitations of perturbation-based post-hoc explanations, 
             such as explanations socially misaligned, OOD in perturbations, algorithmic inefficiencies, 
             etc. The Leaning to Explain (L2X) method based on the information bottleneck theory, which
              is proposed in \cite{bang2021explaining} is extended to more image datasets. We state 
              that imitation-based explanations are unnecessary. Even without approximating a black-box model, 
             it is possible to get meaningful enough feature importances using only raw data labels.  Our main contributions are as follows:

    \begin{itemize}
        \item We propose an explanation-based counterfactual retraining (XCR) method.
        The generation of explanations for retaining needs only one forward pass because XCR is based
        on L2X. The additional computational overhead is low, and the efficiency
         is significantly higher than the existing methods requiring multiple forward passes.
        \item We improve existing imitation-based L2X methods and demonstrate
         that sufficiently compelling features can be extracted in XCR even 
         if not based on imitation of the black-box model.
        \item Our proposed XCR method can effectively improve black-box model
         robustness and OOD calibration performance without modifying the model structure and 
        calibration on the validation set, which experimentally demonstrated on popular vision datasets.
    \end{itemize}

\section{Related work}
\subsection{Explainable Artificial Intelligence}
The research of XAI has attracted much attention recently and is a research hotspot. We can divide 
the related work into semantic-level explanation and mathematical-level explanation. There are many
 studies on explanation methods at the semantic level, such as gradient-based 
 explanation\cite{shrikumar2017learning,simonyan2014deep,smilkov2017smoothgrad,sundararajan2017axiomatic,selvaraju2017grad}, 
 perturbation-based explanation\cite{zeiler2014visualizing,petsiuk2018rise,fong2019understanding,ribeiro2016should}, etc. 
 These methods can be generalized into Feature Importance (FI) explanations. 
 Lundberg et al. propose SHAP\cite{lundberg2017unified}, trying to unify some FI interpretations. 
 Among semantic-level explanations, Jianbo Chen et al. propose the L2X \cite{chen2018learning} 
 method based on information theory, which introduces instance feature selection as a method for the
  black-box model explanation. L2X is based on learning a function to extract the most useful
   subset of features for each given instance. Bang et al. propose VIBI\cite{bang2021explaining}, 
   a model-agnostic explanation method that provides a brief but comprehensive explanation. 
   VIBI is also based on the variational information bottleneck. For each instance, VIBI 
   chooses the key features that compress the input the most (briefness) and provides information 
   about the black-box system's decision on that input (comprehensive). 
   VIBI played a key role in the subsequent Sec. \ref{sec:method1}. \cite{anders2021software,hedstrom2022quantus} propose interactive 
   and operational explanation methods and developed related software packages. 
   \cite{bau2018gan,shen2020interfacegan} introduce methods to explain generative models. 
   With the introduction of the deep neural network model with attention, 
   its attention weights also play an important role in the XAI field\cite{jain2019attention,wiegreffe2019attention,chefer2021transformer}. The explanation at the mathematical level mainly includes explaining the 
   maximum representation or generalization 
   ability\cite{yang2020interpolation,xu2017information,xu2018understanding,shwartz2017opening} of
    the neural network model. Quanshi Zhang et al.\cite{ren2021unified,deng2021discovering,shen2021interpreting,ren2021interpreting} applied 
    game theory to research on explainable artificial intelligence, trying to unify and 
    mathematically prove various existing explanation methods.

  \subsection{OOD Data Challenge in XAI}
  In application scenarios, modern machine learning is challenged by open environments, and anomaly
   detection of OOD data is one of the critical issues. Anomaly detection techniques are widely used
    in machine learning. We can divide these techniques into three categories: supervised, 
    semi-supervised and unsupervised, and the classification depends on the availability of OOD data
     labels\cite{qiu2021resisting}. OOD problems arise when the counterfactual inputs used to 
     create or evaluate explanations are out-of-distribution to the model. In perturbation-based
      methods, the counterfactual data inputs used to generate FI interpretations are considered 
      OOD data because they are different from the training data and contain different features 
      than the training data. \cite{hooker2019benchmark} proposed that the perturbation-based 
      method violates the IID assumption of machine learning. Without retraining, it is unclear 
      whether the model performance decline comes from distribution changes or because the removed
       features can provide rich information. The explanation generation may come from the 
       distribution change of the data rather than the perturbation itself. \\
\indent
    Therefore, Hase et al.\cite{hase2021out} summarized a long line of past work on OOD concerns 
    in XAI, then proposed that explanations generated 
    using a standardly trained neural model are usually
    socially misaligned. 
    %First, \cite{hase2021out} outlines what social expectations
    %are in terms of feature importance explanations, arguing that social expectations are 
     %violated because counterfactuals inputs are OOD. They then propose
     % a method of aligning the training and testing distributions by exposing 
     % the model's counterfactual inputs to the model during the training phase. 
     Therefore, Hase et al.\cite{hase2021out} summarize a long line of past work on OOD
      concerns in XAI, then propose that explanations generated using a standardly trained
       neural model are usually socially misaligned. \cite{hase2021out} propose to carry out
        counterfactual retraining in the field of Natural Language Processing (NLP) and mainly 
        evaluate the model from the perspective of explainability, lacking the exploration of 
        the calibration performance of the model and the vision dataset. \cite{xie2020adversarial}
         propose that adversarial samples can effectively improve the image recognition 
         performance of DNNs. Explanation samples with feature importance, i.e., the counterfactual
          input mentioned in \cite{hase2021out} are very similar to the adversarial samples, 
          and they both modify the original input. We believe that the FI explanation in the 
          counterfactual retraining method can be an adversarial example and extended to the 
          vision dataset. Our approach mainly treats FI explanation as adversarial examples 
          for black-box model retraining. We consider the black-box model's robustness, explainability
           performance, and OOD calibration metrics.

  \section{Method}
  We propose eXplanation-based Counterfactual Retraining (XCR), which improves model robustness
   and OOD calibration capability by adding counterfactual input with feature importance explanation
    while training. We introduce the fast FI-based explanation extraction method 
    based on the information bottleneck principle in Sec. \ref{sec:method1}. The process
     of adding counterfactual samples to black-box model retraining is shown 
     in Sec. \ref{sec:method2}.
  \subsection{Counterfactual Examples Generation}
  \label{sec:method1}
\begin{figure}[htbp]
  \includegraphics[width=\textwidth]{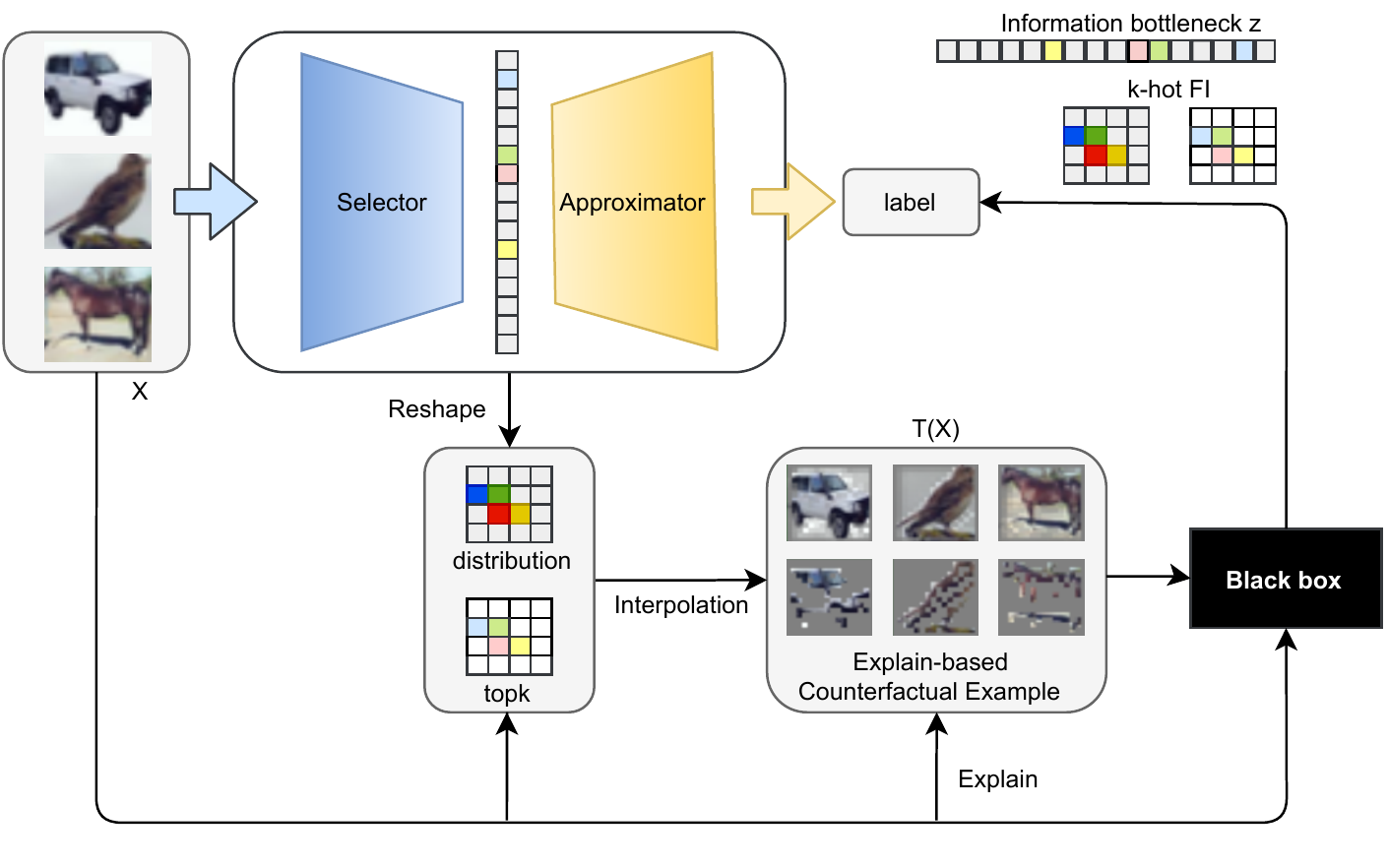}
  \caption{Overall flowchart of XCR. On the one hand, the selector 
  and approximator structure extract $k$-hot feature importance 
  explanations. On the other hand, XCR retains the black-box model by 
  combining the feature importance interpretation of $k$-hot with the original input.} \label{fig1}
\end{figure}

The information bottleneck principle\cite{tishby2000information}
 provides a method to formally and quantitatively describe the performance of a model 
 in learning feature importance using relevant information from the perspective of information theory. 
  Specifically, the optimal model is to transfer as much information as possible from the
   input $\mathbf{x}$ to the output $\mathbf{y}$ by compressing the representation $\mathbf{t}$ (called the information bottleneck). 
   \cite{bang2021explaining} introduces VIBI, a model-agnostic FI interpretation method 
   based on this principle. 
   The only difference between VIBI and L2X\cite{chen2018learning} is that the additional term 
   of VIBI effectively increases the entropy of the distribution $p(\mathbf{z})$, the explanation
    bottleneck. L2X only minimizes the cross-entropy between the prediction of the black-box 
    model and the prediction of the approximator.
   The information bottleneck objective of VIBI optimization is:
   \begin{equation}
    p(\mathbf{z|x}) = \mathop{\arg\max}_{p(\mathbf{z}|\mathbf{x}),p(\mathbf{y}|\mathbf{t})} 
    {\rm I(\mathbf{t},\mathbf{y}) - \beta \ I(\mathbf{x},\mathbf{t}) }
    \end{equation}
    where $\rm I(\cdot,\cdot)$ is the mutual information and 
    $ \rm I(\mathbf{t},\mathbf{y})$ represents the sufficiency of information,
     $-\rm I(\mathbf{x},\mathbf{t})$
    represents the briefness of the explanation $\mathbf{t}$. 
    To balance the relationship between the two, 
    $\beta$ is used as a Lagrange multiplier.\\
\indent
VIBI proposes a method of using an explainer to extract FI explanations 
first and then using an approximator to imitate the behavior of the black-box model. 
That is, the explainer selects the top-$k$ most important features to get the feature 
importance $\rm \mathbf{t} = T(\mathbf{x})$, which provides an instance-specific explanation. 
An image is divided into $n\times n$ smallest units as patches in the vision dataset. 
An image can be divided into $d=n\times n$ patches. Then $\rm T(\mathbf{x})$ 
is used as the approximator input to approximate the behavior of the black-box model. 
    %Explain that $T(\mathbf{x})$ is defined as:
    %\begin{equation}
    %  T(\mathbf{x}) = {(\mathbf{x} \odot \mathbf{z})}
    %\end{equation}
    %where $\mathbf{x}$ is the input and $\mathbf{z}$ is the feature importance generated 
    %by the model. 
    The Variational bound for $\rm I(\mathbf{t},\mathbf{y})$ and $\rm I(\mathbf{x},\mathbf{t})$ is 
    proved similarly as \cite{bang2021explaining} and the total variational bound is obtained:
    \begin{equation}
      \begin{aligned}
       {\rm I(\mathbf{t,y}) }
      & - \beta \ {\rm I(\mathbf{x,t})} \\
      & \ge
      \mathbb{E}_{\mathbf{x}  \sim p(\mathbf{x})}
      \mathbb{E}_{\mathbf{y} | \mathbf{x} \sim p(\mathbf{y|x})}
      \mathbb{E}_{\mathbf{t} | \mathbf{x} \sim p(\mathbf{t|x})}
      [\log q(\mathbf{y|t})] \\
      &- \beta \ \mathbb{E}_{\mathbf{x}  \sim p(\mathbf{x})}
      D_{\rm KL}(p(\mathbf{z|x}),r(\mathbf{z})) + C^*
      \end{aligned}
      \end{equation}
      with proper choices of $p(\mathbf{z|x})$ and $(\mathbf{z})$, 
      we can assume that the Kullback-Leibler divergence
       $D_{\rm KL}(p(\mathbf{z|x}),r(\mathbf{z}))$
      has an analytical form. $C^*$ is independent of the optimization procedure.
      For more proof details, please refer to Supplementary Material S1. \\
      \indent
      Then the \textbf{generalized Gumbel-softmax trick} \cite{jang2016categorical} is used
     to avoid the problem of sampling top-$k$ out of $d$ cognitive patches 
     where each patch is assumed drawn from a categorical distribution. 
     Specifically, we independently sample a cognitive patch for $k$ times, 
     when perturbation $e_j$ is added to $\log$ probability $\log p_j(\mathbf{x})$.
     Then differentiable approximation to $\arg \max$ is defined:
     \begin{equation}
      \begin{aligned}
         g_i &= -\log (-\log e_j), \ \  where \ e_j \sim U(0,1)\\
          c_j  &= \frac{exp((g_j + \log p_j(\mathbf{x}) ) / \tau)} 
          {\sum_{j=1}^d exp((g_j + \log p_j(\mathbf{x})) / \tau)},
      \end{aligned}
      \end{equation}
      where $c_j$ is one element of $\mathbf{c} = (c_1,...,c_d)$ working as a continuous
      and $\tau$ is temperature parameter. There is a continuous-relaxed random vector
      $\mathbf{z^*}= [z_1^*,...z_d^*]^T$ as the element-wise maximum of $\mathbf{c}$:
      \begin{equation}
        z_j^* = \mathop{\max}_l c_j^{(l)} \ , \ \ l = 1,...k 
        \end{equation}
      Then we can use standard backpropagation to compute the gradients.\\
      \indent
        By putting everything together, we obtain:
        \begin{equation}
          \frac{1}{nL} \sum_i^n \sum_l^L \left[ 
            \log q(\mathbf{y_i^{lb}|x_i} \odot f( { \mathbf{e}^{(l)} } , \mathbf{x_i}))
            - \beta \ D_{\rm KL} (p(\mathbf{z^*|x_i}),r(\mathbf{z^*}))
           \right]
        \end{equation}
        where $q(\mathbf{y_i^{lb}|x_i} \odot f( { \mathbf{e}^{(l)} } , \mathbf{x_i}))$ aims to 
        fit the origin label $\mathbf{y_i^{lb}}$, but not to approximate the black box model. 
        $ D_{\rm KL} (p(\mathbf{z^*|x_i}),r(\mathbf{z^*}))$ ndicates how well the information
         is compressed. \\
        \indent
        The VIBI explanation method works by only imitating the output of the black-box model, but we find 
        that this behavior may be unnecessary. In short, the performance of the current state-of-art 
        black-box model in most tasks is good enough. For example, the image classification accuracy
         of CIFAR10 in \cite{kolesnikov2020big,dosovitskiy2020image} is over 99\%, 
         in CIFAR100\cite{foret2020sharpness} is over 95\%, and in 
         Imagenet1k\cite{wortsman2022model,dai2021coatnet} is over 90\%. 
         Therefore, there is no significant difference between approximating
        the output of the black-box model and learning the original labels directly.
        The experiment Sec. \ref{sec:ex1} confirms our view that it is not vital which
         existing black-box model to imitate in the post-hoc explanation. 
         The key to the problem is not to explain the existing black-box model 
         but to extract useful features. Even if the black-box model is not required, 
         the extracted features are similar to or outperform the results of the approximation
          of the black box model in VIBI.
      Therefore, as shown in Fig. \ref{fig1}, the original black-box model output $\mathbf{\hat y}$ is replaced with the original label $\mathbf{y}$.
      To make the method more suitable for RGB image datasets 
      instead of MNIST datasets and natural language datasets, 
      in addition to using $\mathbf{Topk}$ to select explanations, 
      we also use a $\mathbf{Softmax}$ based method to select explanations.
      The formula of converting $\mathbf{z}$ to $T(\mathbf{x})$ is:
      \begin{equation}
        \rm T(\mathbf{x})
        = \mathbf{Interpolate}(\mathbf{Reshape}(\mathbf{S}(\mathbf{z}, dim = 1))) \odot \mathbf{x}
      \end{equation}
where $\mathbf{Interpolate}$ is the interpolation function that can convert a $(B, C, h ,w)$ 
tensor to the shape of the original image. The Reshape function transforms 
a tensor of shape $(B, C \times h \times w)$ into $(B, C, h, w)$. 
$S$ is the selection function, which can be $\mathbf{Softmax}$ or $\mathbf{Topk}$,
 and is used to select the most important explanations.
  \subsection{Black-Box Model Retrain}
  \label{sec:method2}
  As shown in Fig. \ref{fig1}, We propose an explanation-based retraining pipeline. 
  Specifically, as discussed in Sec. \ref{sec:method1}, existing 
  L2X methods\cite{chen2018learning,bang2021explaining} describe the behavior of black-box models
   by approximating them. However, this explanation process may have little to do with
    the black-box model's internal structure but much to do with the structure and parameters
     of the L2X model. We strongly believe that this is similar to the situation described 
     in \cite{hase2021out} and is also a social misalignment in explanation. The human expectation
      is that feature importance explanations can reflect how models learn to use feature
       explanations as evidence for a particular decision. However, the presence of OOD affects
        FI explanations with model hyperparameters, random seeds of initialization, and data
         ordering\cite{probst2019tunability,dodge2020fine,van2017empirical}. In L2X, the 
         approximator also uses the FI explanation data 
  to approximate the black-box model, so naturally, there is a social misalignment.
  In addition, approximating black-box models can also introduce social misalignment. 
  Specifically, the social expectation during the training process is to explain the
   black-box model. However, the L2X method does not pay attention to details 
  of the black-box model but only to the input and output of the black-box model.\\
  \indent
      Therefore, we propose to use the original labels to generate feature 
      importance explanations without approximating the black-box model, 
      and then use the generated feature importance
       explanations as counterfactual input to join the black-box model 
      for retraining.

  \section{Experiment and Results}
  We carry out experiments on popular vision datasets. 
  We describe evaluation metrics of XAI, robustness, and OOD calibration
   in Sec. \ref{sec:method3}. We state the unnecessary use of imitating the behavior 
   of the black-box model in L2X  in Sec. \ref{sec:ex1}. We explain the explanation-based
    counterfactual training process of the black-box model in detail and evaluate the XCR method in Sec. \ref{sec:ex2}. 

  %To confirm that the method is also 
   %applicable in transfer learning, experiments with 
  %counterfactual samples involved in fine-tuning are described in \ref{sec:ex3}.

  \subsection{Objective Evaluation Metrics}
  \label{sec:method3}
  We evaluate the XCR method in both the explanation and OOD 
  calibration to compare it with existing methods.
  \subsubsection{XAI Evaluation Metrics}
  As far as we know, there is no generally accepted evaluation system for the XAI model. 
  Therefore, in order to ensure the credibility of the results, we selected the evaluation indicators used in
   \cite{hase2021out,wang2020score,chattopadhay2018grad} for comparison. The explanation $e$ is described in \cite{hase2021out} 
   as a $k$-sparse binary vector in ${\{0, 1\}}^d$, where $d$ is the dimensionality of the feature space.
   The explanation $e$ is similar to the $k$-hot vector encoding the information bottleneck in Sec. \ref{sec:method2}.
   However, in image classification tasks, $k$-hot vectors need to be reshaped and upsampled to fit the image size.
    For models $f$ and information bottleneck $z$ where the model output is the class classification probability, 
    $\mathbf{Sufficiency}$ can be defined as:
    \begin{equation}
      \mathbf{Suff}(f,\mathbf{x},\mathbf{z},s) = f(x)_{\hat y} - f(\rm T_s(\mathbf{x},\mathbf{z}))_{\hat y} 
    \end{equation}
  where $\hat{y} = \arg \max_y f(x)_y$ is the predicted class, $T_s(\mathbf{x},\mathbf{z})$ is feature importance 
  explanations extracted from information bottlenecks, $s$ is the proportion of information retained.

   We also follow the \textbf{Average Drop} and \textbf{Average Increase} 
   in \cite{chattopadhay2018grad}.
   The \textbf{Average Drop} is expressed 
   as $\sum_{i=1}^N \frac{\max(0,Y_i^c - E_i^c)}{Y_i^c}$.
  The \textbf{Average Increase} is expressed as
    $\sum_{i=1}^N \frac{Sign(Y_i^c < E_i^c)}{N}$.
    Where $Y_i^c$ is the predicated probability for class $c$ on image $i$ and $E_i^c$ is the 
    predicated probability for class $c$ with k-hot explanation map region image. $Sign$ will return $1$ if input is $True$.
   The \textbf{Fidelity} is the consistency of the FI-explained sample and the output of the 
   original sample input into the model.

  \subsubsection{Model Performance and OOD Calibration Metrics}
  The performance of the original approximator can be used as a reference but is not decisive because the 
  black-box model needs to be retrained later. The structure of the original selector and 
  approximator may be simple, and the extracted vital feature information is only a tiny proportion.
   Thus, there will be performance loss like most XAI methods. 
   We mainly use the image classification accuracy of the retrained model on the 
   (corrupted-)vision dataset to evaluate model performance.\\
   \indent
  To evaluate the robustness and OOD calibration performance of the model after the XCR 
  method is used, we use the benchmark proposed in \cite{hendrycks2018benchmarking}. 
  Negative Log-Likelihood (\textbf{NLL}\cite{hastie2009elements}), Expected Calibration Error (\textbf{ECE}\cite{naeini2015obtaining}) 
  and \textbf{Brier}\cite{brier1950verification}
  are also used to compare with the existing methods \cite{tian2021geometric}.

  \begin{table}[htbp]
    \centering
    \caption{Explanation metrics across methods, parameters, and 
    datasets. We compare feature extraction with different
     retain ratios and different datasets (CIFAR10/100) when $patch size = 2$. 
     Moreover, we report the evaluation metrics at each $\beta$ value. 
     Accuracy and explanation metrics \textbf{Average Drop} 
    and \textbf{Average Increase} are used to compare VIBI with our method.
      }\label{tab1}
    %\begin{tabular}{c|c|c|C{1.5cm}C{1.5cm}|C{1.5cm}C{1.5cm}|C{1.5cm}C{1.5cm}}
      \begin{tabular}{c|C{0.8cm}|C{1cm}|C{1.3cm}C{1.3cm}|C{1.3cm}C{1.3cm}|C{1.3cm}C{1.3cm}}
      \hline
    \multirow{2}{*}{Dataset} & \multirow{2}{*}{k}  & \multirow{2}{*}{$\beta$}  
    & \multicolumn{2}{c|}{Accuracy $\uparrow$} 
    & \multicolumn{2}{c|}{Avg Drop $\downarrow$} 
    & \multicolumn{2}{c}{Ave Inc $\uparrow$} \\
    \cline{4-9}
        &  & & VIBI & XCR & VIBI & XCR& VIBI & XCR\\
    \hline
    \multirow{9}{*}{CIFAR10} 
    & 16 & 0.1	  &\textbf{0.649}&	0.550 & 0.717 &\textbf{0.359} &0.050 & \textbf{0.117}\\
    & 16 & 0.01	  &\textbf{0.675}&	0.618 & 0.825 &\textbf{0.362} &0.039 & \textbf{0.092}\\
    & 16 & 0.001	&\textbf{0.686} & 0.648 & 0.851 & \textbf{0.533} &0.042 &  \textbf{0.061}\\

    & 32 & 0.1	  &0.737&	\textbf{0.765} & 0.192 &\textbf{0.119} &0.166 & \textbf{0.219}\\
    & 32 & 0.01	&0.334&	\textbf{0.791} & \textbf{0.065} &0.539 &\textbf{0.341} & 0.057\\
    & 32 & 0.001	&\textbf{0.758} &0.751  & 0.670 &\textbf{0.143} &0.045 & \textbf{0.182}\\

    & 64 & 0.1	  &0.786&	\textbf{0.828} & 0.090 &\textbf{0.015} &0.287 & \textbf{0.450}\\
    & 64 & 0.01	&0.785&	\textbf{0.856} & \textbf{0.232} & 0.249 &\textbf{0.166} &0.154\\
    & 64 & 0.001	&0.796 &\textbf{0.847}  & \textbf{0.200} & 0.208&\textbf{0.183} & 0.159\\
    \hline

    \multirow{4}{*}{CIFAR100}

    %& 32 & 0.1  & &	 &  & & & \\
    & 32 & 0.01	&0.418&	\textbf{0.449} & 0.455 &\textbf{0.193} &0.086 & \textbf{0.191}\\
    & 32 & 0.001	&0.449 &\textbf{0.483}  & \textbf{0.510} &0.642 &\textbf{0.061} & 0.032\\

    %& 64 & 0.1	  & &	 &  & & & \\  
    & 64 & 0.01	&0.496&	\textbf{0.567} & \textbf{0.099} & 0.104 &0.261 &\textbf{0.265}\\
    & 64 & 0.001	&0.496 &\textbf{0.574}  & 0.417 & \textbf{0.169} &0.101 & \textbf{0.199}\\

    \hline
  \end{tabular}
    \end{table} 

  \subsection{Learning to Explain without Imitation}
  \label{sec:ex1}

\begin{figure}[!htbp]
          \includegraphics[width=\textwidth]{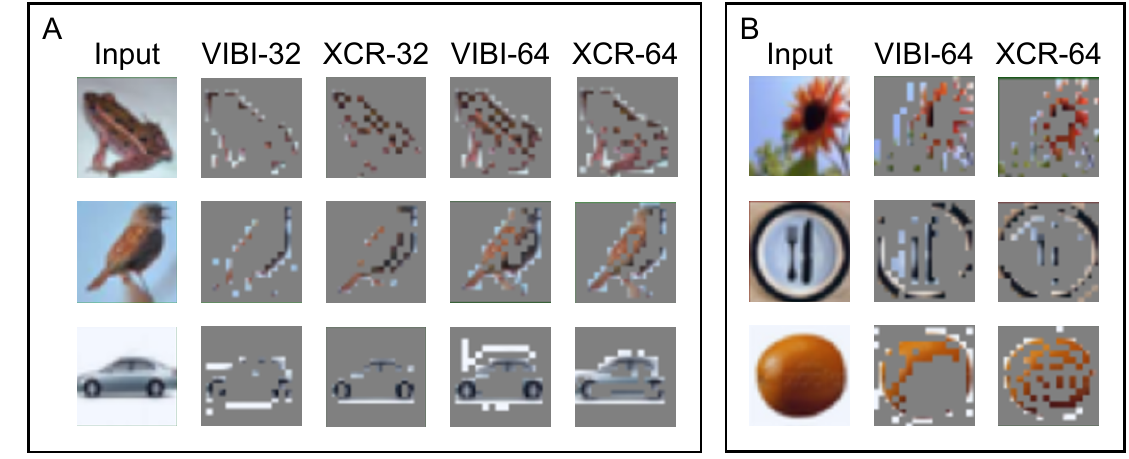}
          \caption{
          VIBI based on imitating the black-box model and our feature extraction without
           imitation. A. CIFAR10 dataset, we select the top-$32$ 
           and top-$64$ explanation patches to show. 
           B. CIFAR100 dataset, we select the top-$64$ explanation patches to show.} \label{fig2}
        \end{figure}

  %Our experiments in this part aim to show that there is no significant
   %difference between imitating the black-box model and using the original
    %labels directly. 
Based on the VIBI method described in \cite{bang2021explaining}, 
we change the imitation behavior of the approximator to fitting the original
 labels, thereby ignoring the influence of the black-box model. We use \textbf{ResNet28-10} 
 trained according to \cite{zagoruyko2016wide} as the black-box model output for the L2X 
 approximation. In the explained structure, \textbf{ResNet18} is used as a selector, 
 and \textbf{ResNets}\cite{he2016identity} of various structures are used as approximators
  to adapt to different hyperparameter requirements. We use the \textbf{CIFAR10/100} datasets
   and search for parameters 
   $patch size=\{2,4\}$, $k=\{8,16,32,64\}$ and $\beta=\{1,0.1,0.01,0.001\}$, part of 
   the experimental results of \textbf{CIFAR10/100} are shown in Table \ref{tab1}. 
   More intuitively, we show the image's explanation in Fig. \ref{fig2}. 
Please refer to Supplementary Materials S2 and S3 for more detailed results.\\
\indent From the explanation results for different parameters and datasets, 
there is no significant difference between VIBI and our XCR method. With $k = 16$ and a 
low feature retaining rate of $6.25\%$, VIBI's approximator has slightly better 
accuracy than our XCR method, but VIBI's XAI metric performs poorly. At regular feature 
retention rates of $12.5\%$ and $25\%$, our XCR method outperforms the original VIBI 
method on most metrics. Fig. \ref{fig2} also visually shows that our XCR can generate 
meaningful results under the same parameter settings, and there is no significant difference 
from imitating the black-box model. 
Even the XCR explanations of some images make more sense than VIBI intuitively.
      
  \subsection{Counterfactual Retraining}
  \label{sec:ex2}

  As discussed in Sec. \ref{sec:ex1}, the original label can be used directly
   without imitating the black-box model. We use the standard-trained \textbf{ResNet28-10}
    as the black-box model without using additional data in \cite{zagoruyko2016wide}, 
    which is convenient to compare the results in \cite{tian2021geometric}. 
    In order to simplify the experiment, we use the same simple selectors 
    and approximators as Sec. \ref{sec:ex1} with not too many parameters. 
    We do not rule out the possibility of using more complex selectors 
    and approximators to get better results. We use the \textbf{CIFAR10/100}
     datasets and the benchmark proposed in \cite{hendrycks2018benchmarking},
      with the corruption image datasets \textbf{CIFAR10-C}, \textbf{CIFAR100-C} 
      with 16 types of noises with 5 
      severity scales. The robustness and performance improvement of OOD calibration brought 
      by the XCR method is verified.

\subsubsection{Parameter Search in Retraining}
When retraining, we set the learning rate to 0.1 according to the
 training process set in \cite{zagoruyko2016wide}, the depth of the \textbf{Wide-ResNet} is set to $28$, 
 the widen factor is set to $10$, and dropout is not used. The vanilla model was trained for 200 
 epochs as a comparison. Similar to Sec. \ref{sec:ex1}, we do not modify the vanilla model and
  use a consistent model structure and the same training process, 
  but only add counterfactual samples. Through pre-experiments, we find that XCR performs
   best when the patch size is 2.
  We perform a grid search on the XCR parameters, including $k={16,32,64}$
   and $\beta={0.1,0.01,0.001}$. The results compared with existing methods are shown in Table \ref{tab2}, and the 
   specific results of parameter grid search are shown in Supplementary Material S2.
   Our XCR method when $k=32$,$\beta=0.001$ beats all methods in accuracy on corrupted data, 
    and also outperforms most existing methods such as single-pass method 
    \textbf{DUQ}\cite{van2020uncertainty}, \textbf{SNGP}\cite{liu2020simple} and even multi-pass
     method \textbf{Deep Ensembles}\cite{lakshminarayanan2017simple},
     \textbf{MC Dropout}\cite{gal2016dropout}, \textbf{GSD}\cite{tian2021geometric} in \textbf{NLL}, \textbf{ECE}.
   
   \begin{table}
    \centering
    \caption{Calibration metrics across methods and datasets. The experimental 
    results are averaged with 10 random seeds. $\dagger$ represents the result 
    in \cite{tian2021geometric}. Our method when $k=32$, $\beta=0.001$ beats 
    all methods in accuracy on corrupted data, 
    and also outperforms most existing methods in \textbf{NLL}, \textbf{ECE}.}\label{tab2}
    \begin{tabular}{c|l|C{1.2cm}C{1.2cm}|C{1.2cm}C{1.2cm}|C{1.2cm}C{1.2cm}}
    \hline
    \multirow{2}{*}{Dataset} & \multirow{2}{*}{Method} 
    & \multicolumn{2}{c|}{Accuracy $\uparrow$} 
    & \multicolumn{2}{c|}{ECE $\downarrow$} 
    & \multicolumn{2}{c}{NLL $\downarrow$} \\
    \cline{3-8}
        &  & Clean & Corrupt & Clean &  Corrupt  & Clean & Corrupt \\
    \hline
    \multirow{7}{*}{CIFAR10} 
    & Vanilla{$\dagger$} &96.0 &	72.9  & 0.023 &0.153 & 0.158 & 1.059\\
    & DUQ{$\dagger$}  &94.7 &	71.6  & 0.034 &0.183 & 0.239 & 1.348\\
    & SNGP{$\dagger$}	&95.9 &	74.6  & 0.018 &0.090 & 0.138 & 0.935\\

    & Deep Ensembles{$\dagger$}	 &96.6 &	77.9  & 0.010 &0.087 & 0.114 & 0.815\\
    & MC Dropout{$\dagger$} &96.0 &	70.0  & 0.021 &0.116 & 0.173 & 1.152\\
    & GSD{$\dagger$}	&\textbf{96.6} &	77.9  & 0.007  & 0.069 & 
    \textbf{0.108} & 0.773 \\

    & Ours XCR	 & 96.3    &	\textbf{80.6}  & \textbf{0.002} & \textbf{0.062}
     & 0.122     & \textbf{0.711}         \\
    \hline

    \multirow{7}{*}{CIFAR100} 

    & Vanilla{$\dagger$} &79.8 &	50.5  & 0.085 &0.239 & 0.872 & 2.756\\
    & DUQ{$\dagger$} &78.5 &	50.4  & 0.119 &0.281 & 0.980 & 2.841\\
    & SNGP{$\dagger$}	&79.9 &	49.0  & 0.025 &0.117 & 0.847 & 2.626\\

    & Deep Ensembles{$\dagger$}	 &80.2 &	54.1  &0.021 &0.138 & 0.666 & 2.281\\
    & MC Dropout{$\dagger$}	&79.6 &	42.6  & 0.050 &0.202 & 0.825 & 2.881\\
    & GSD{$\dagger$}	&83.0 &	54.1  & 0.018  & \textbf{0.086} & 0.614 & \textbf{2.042} \\

    & Ours XCR	 &  \textbf{85.0}  &	\textbf{56.5}  & \textbf{0.002}  & 0.095  & \textbf{0.530} & 2.113        \\

    \hline
    \end{tabular}
    \end{table}

\begin{figure}[!htbp]
      \includegraphics[width=\textwidth]{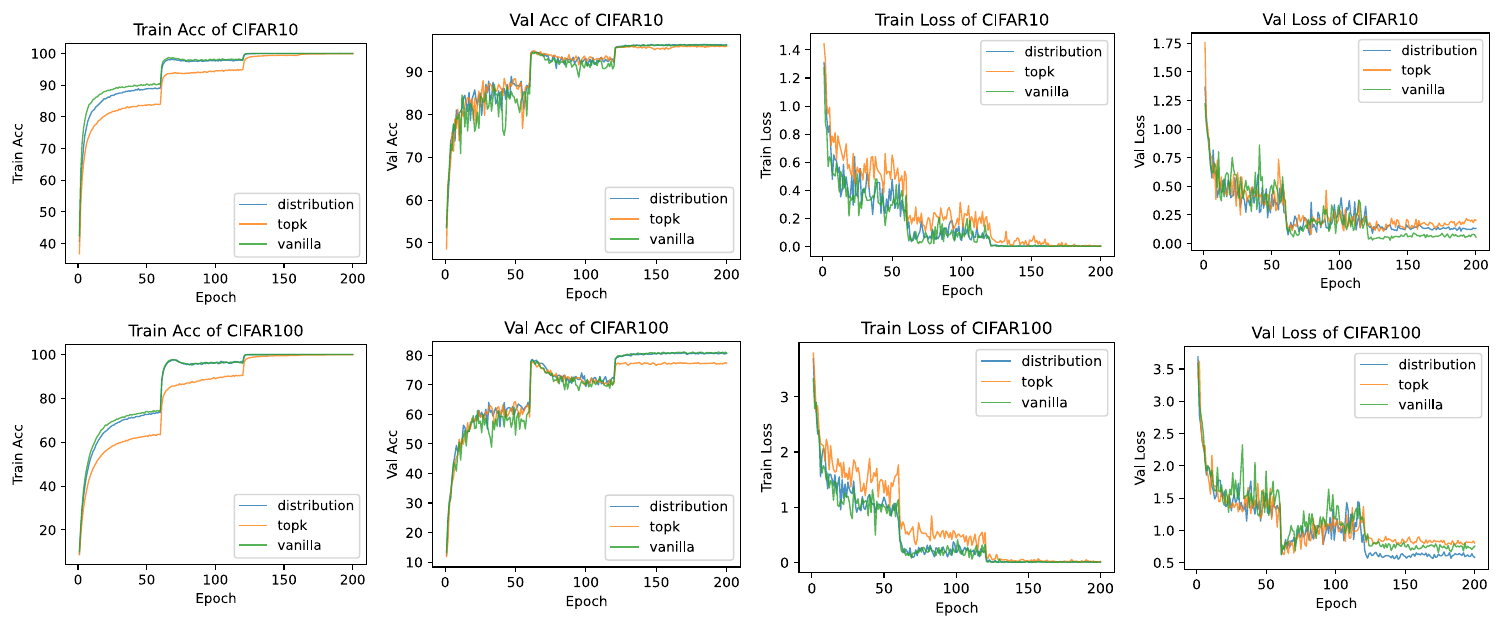}
      \caption{
        Convergence curves and accuracy curves when $k=32$, $\beta = 0.001$.} \label{fig3}
    \end{figure}

    \indent
    The training process when $k=32$, $\beta = 0.001$ is shown in Fig. \ref{fig3}.
    We plot the metrics of the training set and the validation set 
    explained by $\mathbf{Topk}$ and $\mathbf{Softmax}$ during 
    the training process and compare them with vanilla results.
    It can be seen from the training process of 200 rounds that when we use the distribution
     explanation mode, that is, when $\mathbf{Softmax}$ is selected as the map function of
      processing the information bottleneck, the convergence of the model is better 
      than $\mathbf{Topk}$. Our XCR method improves the robustness of the black-box model 
      without significantly different from vanilla convergence speed. The validation loss 
      of the XCR method on a dataset with a small number of categories is slightly higher 
      than that of the vanilla method. However, the effect is not apparent, and the validation 
    loss on a dataset with many categories is lower than that of the vanilla method.
    
    \subsubsection{Model Performance without Temperature Calibration}
    In the existing calibration methods, the temperature parameters of the DNNs head need to 
    be calibrated on the validation set, such as \cite{tian2021geometric}. Access to the 
    validation set is limited to many problems, so we hope to achieve satisfactory calibration 
    results without using the validation set calibration. We perform Our experiments on 
    GSD\cite{tian2021geometric} and XCR methods without validation set calibration, then the
     OOD calibration metrics and model accuracy are compared. The results are shown in Table
      \ref{tab3}. We reproduce the experimental results using the same parameter settings 
      of the GSD\cite{tian2021geometric} open source code and add experiments without calibration 
      on the validation set. We find that the uncalibrated GSD method performs worse than our 
      XCR method. Our XCR method, even without validation set calibration, exhibits satisfactory
       image classification accuracy on the corrupted dataset. When evaluating without validation
        set temperature calibration, \textbf{ECE}, \textbf{NLL} and \textbf{Brier} metrics are 
        not significantly different between GSD and our XCR method.

    \begin{table}
      \centering
      \caption{Calibration metrics when validation set is not used.
      The experimental results are averaged with 10 random seeds. 
      Our method when $k=32$, $\beta=0.001$ 
        exhibits satisfactory image classification accuracy on the corrupted dataset. 
        \textbf{ECE}, \textbf{NLL} and \textbf{Brier} metrics are
        not significantly different between GSD and our XCR method.} 
      
      \label{tab3}
    \begin{tabular}{cc|C{1cm}C{1cm}C{1cm}C{1cm}|C{1cm}C{1cm}C{1cm}C{1cm}}
      \hline
      \multirow{2}{*}{} & \multirow{2}{*}{} 
      & \multicolumn{4}{c|}{CIFAR10} 
      & \multicolumn{4}{c}{CIFAR100}  \\
      \cline{3-10}
           &        & Acc $\uparrow$  & ECE $\downarrow$   & NLL $\downarrow$   & Brier$\downarrow$ 
           & Acc $\uparrow$  & ECE $\downarrow$   & NLL $\downarrow$   & Brier$\downarrow$ \\
      \hline
      \multirow{2}{*}{Clean} 
      & GSD         &96.28  &   0.013 & 0.147 & 0.006 & 79.74  & 0.018  &  0.815   &  0.003     \\
      & XCR         &\textbf{96.61}  &  0.015 & 0.153 & 0.006 & \textbf{80.98}& 0.034 & 0.773   & 0.003 \\
      \hline
  
      \multirow{2}{*}{Corrupted} 
      & GSD        &69.33  & 0.091 & 0.838 & 0.035            &  48.75       &  0.118    &  2.315       & 0.007\\
      & XCR         &\textbf{80.25}  & 0.104 & 0.841 & 0.031  & \textbf{58.20} & 0.120 & 2.148 & 0.006 \\
  
      \hline
      \end{tabular}
      \end{table}

\subsubsection{Explanation Performance after Retraining}
Before retraining, as shown by the evaluation metrics in Table \ref{tab1}, 
the explanation performance of the model is not good. After applying our XCR method, 
the model's explanation fidelity, Average Drop,  Average Increase, and Sufficiency are
 all significantly improved to a satisfactory 
level. The results can be found in Table \ref{tab4}.
For more details, please refer to Supplementary Material S2.

\begin{table}
  \centering
  \caption{Explanation metrics when XCR is used.
  To illustrate that XCR can maintain high explainability
    while calibrating the model, we evaluate the retrained 
    model using \textbf{Fidelity}, 
    \textbf{Average Drop}, \textbf{Average Increase}, 
    and \textbf{Sufficiency} and find that it is significantly 
    better than that in Sec. \ref{sec:ex1} result.
    }\label{tab4}
  %\begin{tabular}{c|c|c|C{1.5cm}C{1.5cm}C{1.5cm}C{1.5cm}}
    \begin{tabular}{c|C{1cm}|C{1.7cm}C{1.7cm}C{1.7cm}C{1.7cm}}
  \hline
  {Dataset} & {k}  
  %& {$\beta$}  
  & {Fidelity $\uparrow$} 
  & {Avg Drop $\downarrow$} 
  & {Avg Inc $\uparrow$}
  & {Suff $\downarrow$} \\
  \hline
  \multirow{3}{*}{CIFAR10} 
  % &16	&0.1		&0.981	&		0.018	&	\textbf{0.501}	&		0.010\\
  %&16	&0.01	&	0.978	&		0.029	&		0.484	&		0.019\\
  %&16	&0.001	&	0.984	&		0.016	&		0.497	&		0.008\\
  %&32	&0.1	&	0.973		&	0.028	&		0.463	&		0.019\\
  %&32	&0.01	&	0.972		&	0.029	&		0.491	&		0.018\\
  %&32	&0.001&		0.971	&		0.033	&		0.483	&		0.022\\
  %&64	&0.1	&	0.974		&	0.026		&	0.447		&	0.016\\
  %& 64	&0.01	&	0.977		&	0.026		&	0.455		&	0.016\\
  %&64	&0.001	&	\textbf{0.985}	&		\textbf{0.015}	&		0.438	&		\textbf{0.008}\\
  &16		&	0.984	&		0.016	&		\textbf{0.497}	&		0.008\\
  &32	&	0.972		&	0.029	&		0.491	&		0.018\\
  &64		&	\textbf{0.985}	&		\textbf{0.015}	&		0.438	&		\textbf{0.008}\\
  \hline

  \multirow{2}{*}{CIFAR100}

  %&32&	0.01	&	\textbf{0.889}	&		0.105	&		0.504		&	0.026\\
  %& 32&	0.001	&	0.880	&		0.108	&		\textbf{0.549}	&		\textbf{0.020}\\
  %&64	&0.01	&	0.887		&	\textbf{0.100}	&		0.534	&		0.021\\
  & 32	&	0.880	&		0.108	&		0.549	&		\textbf{0.020}\\
  &64	&	\textbf{0.887}		&	\textbf{0.100}	&		\textbf{0.534}	&		0.021\\

  \hline
  \end{tabular}
  \end{table}

\subsubsection{Compare with Random Result}
\cite{hooker2019benchmark} shows that many methods are inferior to random mask results, 
so we generate random $k$-hot importance vectors to add to retraining
and compare with XCR results. The results are shown in Supplementary Material S2, 
indicating that the FI-based XCR method outperforms the random mask method both on
$\mathbf{Topk}$ and $\mathbf{Softmax}$ method.

 % \subsection{Counterfactual Input in Fine-tuning of Transfer Learning}  
 % \label{sec:ex3}

  \section{Conclusion}
  In this paper, we argue that the imitation of the black-box model can
   be omitted in existing post-hoc L2X exploration methods that only care about
    the input and output of the model. We then propose an eXplanation-based 
    Counterfactual Retraining (XCR) method to calibrate the OOD problem in XAI.
     XCR can be used to improve the model's robustness and OOD calibration ability. 
     Because the validation set is inaccessible in most cases, we also show that XCR 
     outperforms existing methods without using the validation set for calibration. 
     After retraining, the XAI evaluation metrics of the black-box model will enhance significantly.

%\subsubsection{Acknowledgements} Please place your acknowledgments at
%the end of the paper, preceded by an unnumbered run-in heading (i.e.
%3rd-level heading).

\bibliographystyle{splncs04}
\bibliography{mybibliography}
%Author, F.: Article title. Journal \textbf{2}(5), 99--110 (2016)

%Author, F., Author, S.: Title of a proceedings paper. In: Editor,
%F., Editor, S. (eds.) CONFERENCE 2016, LNCS, vol. 9999, pp. 1--13.
%Springer, Heidelberg (2016). \doi{10.10007/1234567890}

%Author, F., Author, S., Author, T.: Book title. 2nd edn. Publisher,

%Author, A.-B.: Contribution title. In: 9th International Proceedings
%on Proceedings, pp. 1--2. Publisher, Location (2010)

%LNCS Homepage, \url{http://www.springer.com/lncs}. Last accessed 4
%\end{thebibliography}
\end{document}